\documentclass{midl} % Include author names
\usepackage{multirow}
\usepackage{array}
\usepackage{hyperref}
\newcommand{\ie}{\textit{i}.\textit{e}. }
\newcommand{\eg}{\textit{e}.\textit{g}., }
\usepackage{mwe} % to get dummy images
\jmlrvolume{-- XXXX}
\jmlryear{2019}
\jmlrworkshop{Full Paper -- MIDL 2019}
\title[XLSor: A Robust and Accurate Lung Segmentor on Chest X-Rays]{XLSor: A Robust and Accurate Lung Segmentor on Chest X-Rays Using Criss-Cross Attention and Customized Radiorealistic Abnormalities Generation}

\midlauthor{\Name{You-Bao Tang\midljointauthortext{Contributed equally}\nametag{$^{1}$}} \Email{youbao.tang@nih.gov}\\
\Name{Yu-Xing Tang\midlotherjointauthor\nametag{$^{1}$}} \Email{yuxing.tang@nih.gov}\\
\Name{Jing Xiao\nametag{$^{2}$}} \Email{xiaojing661@pingan.com.cn}\\
\Name{Ronald M. Summers\nametag{$^{1}$}} \Email{rms@nih.gov}\\
\addr $^{1}$ Imaging Biomarkers and Computer-Aided Diagnosis Laboratory, Radiology and Imaging Sciences, National Institutes of Health Clinical Center, Bethesda, MD 20892-1182, USA \\
\addr $^{2}$ Ping An Insurance Company of China, Shenzhen, 510852, China
}

\begin{document}

\maketitle

\begin{abstract}
This paper proposes a novel framework for lung segmentation in chest X-rays. It consists of two key contributions, a criss-cross attention based segmentation network and radiorealistic chest X-ray image synthesis (\ie a synthesized radiograph that appears anatomically realistic) for data augmentation.
The criss-cross attention modules capture rich global contextual information in both horizontal and vertical directions for all the pixels thus facilitating accurate lung segmentation. 
To reduce the manual annotation burden and to train a robust lung segmentor that can be adapted to pathological lungs with hazy lung boundaries, an image-to-image translation module is employed to synthesize radiorealistic abnormal CXRs from the source of normal ones for data augmentation. 
The lung masks of synthetic abnormal CXRs are propagated from the segmentation results of their normal counterparts, and then serve as pseudo masks for robust segmentor training. In addition, we annotate 100 CXRs with lung masks on a more challenging NIH Chest X-ray dataset containing both posterioranterior and anteroposterior views for evaluation. Extensive experiments validate the robustness and effectiveness of the proposed framework. The code and data can be found from \url{https://github.com/rsummers11/CADLab/tree/master/Lung_Segmentation_XLSor}.
\end{abstract}
	

\section{Introduction}
\label{intro}
Lung diseases and disorders are one of the leading causes of death and hospitalization throughout the world. According to the American Lung Association, lung cancer is the number one cancer killer of both women and men in the United States, and more than 33 million Americans are facing a chronic lung disease. The chest radiograph (chest X-ray, or CXR) is one of the most requested radiologic examination for pulmonary diseases such as lung cancer, chronic obstructive pulmonary disease (COPD), pneumonia, tuberculosis, etc. There are huge demands on developing computer-aided diagnosis/detection (CADx/CADe)  methods to assist radiologists and other physicians in reading and comprehending chest X-ray images~\cite{Shin_2016_CVPR, Wang_cvpr17, Wang_cvpr18, Tang_MLMI}, given the fact that there is a shortage of experienced radiologists, especially in developing countries.
Precise segmentation of lung fields can provide rich structural information such as shape irregularity, size measurement and total lung volume, which further facilitates subsequent stages of automated diagnosis (\eg disease pattern recognition, segmentation and quantization) to assess certain serious clinical conditions. % In addition, explicit lung masks improve the interpretability of 

Over the past decades, automated segmentation of lung boundaries in CXR has received substantial attention in the literature~\cite{Candemir_TMI14, dai17scan} but still remained a challenging problem~\cite{ el2016biomedical}. Previous work mainly adopted hand-crafted features to design rule-based systems~\cite{LI2001629}, active shape/appearance models~\cite{XU2012452}, or their hybrid methods~\cite{Candemir_TMI14} to segment the lung boundaries. These approaches rely on the test CXR images being well modeled by the existing training images but they may fail on a different distribution or population. Recently, deep learning based methods (e.g. fully convolutional neural networks (FCN)~\cite{FCN}) have achieved great successes in biomedical image segmentation~\cite{DRINet,tang2019ct,cai2018accurate,tang2018ct} and other medical image analysis tasks \cite{tang2019deep,tang2019abnormal,tang2019uldor,tang2018semi,jin2018ct,yan2018deep,yan2019fine}. 
The FCN-based methods are intrinsically limited to local receptive fields and insufficient contextual information due to the fixed geometric structures of the convolution. These limitations impose unfavorable effects in segmenting boundaries around less clear lung regions caused by pathological conditions or poor image quality (\eg low contrast, costophrenic angle clipped off, bad positioning of the patient). Structure correcting adversarial network (SCAN)~\cite{dai17scan} incorporates FCN and adversarial learning~\cite{GAN} to segment organs (lungs and heart) in CXRs. SCAN imposes regularization based on the physiological (global) structures by using a critic network that discriminates between the ground truth annotations from the segmentation masks generated by the FCN.

In order to capture richer global contextual information for robust and accurate lung segmentation, we make use of a criss-cross attention (CCA) module~\cite{Huang2018CCNetCA} to aggregate long-range pixel-wise contextual information in both horizontal and vertical directions. Further dense contextual information can be achieved by stacking more CCA modules recurrently to cover all the pixels. In addition, since publicly available datasets only contain small numbers of lung masks and they are mainly for normal lungs and lungs with subtle findings or unique pathology in an posterioranterior view (\eg small nodules within the lung field in the JSRT database~\cite{JSRT}, CXRs with tuberculosis presented in the Montgomery database~\cite{TB_data}), it is insufficient to directly use these datasets for training a powerful lung segmentor that can be adapted to pathological lungs with hazy lung boundaries (\eg  large masses, pneumonias, effusions, etc.) for both posterioranterior (PA) and anteroposterior (AP) views. Furthermore, it is very time consuming and tedious for radiologists to manually annotate lung masks, especially on CXRs with abnormalities/pathologies in lung regions (or the so-called abnormal CXRs in this paper). Therefore, we use an image-to-image translation method~\cite{munit} to synthesize radiorealistic (\ie a synthesized radiograph that appears anatomically realistic) abnormal CXRs from the source of normal ones for data augmentation and mask propagation. The lung masks of synthetic abnormal CXRs are transferred from their normal counterpart and then used as pseudo masks for segmentor retraining.

The proposed framework \textbf{XLSor} (\ie \textbf{X}-ray \textbf{L}ung \textbf{S}egment\textbf{or}) takes advantage of radiorealistic synthesized abnormal CXRs and pseudo masks, without requiring paired normal and abnormal CXRs from the same patient (which is infeasible in reality), as well as the criss-cross attention module to generate robust and accurate lung segmentation. We annotate 100 lung masks on a more challenging NIH Chest X-ray dataset~\cite{Wang_cvpr17} containing both PA and AP views for evaluation. Extensive experiments on different datasets validate the robustness and effectiveness of the proposed framework.

\section{Methodology}

\subsection{XLSor Framework Overview}
\label{xlsor}
%%%%%%%%%%%%%%%%%%%%%%%%%%%%%%%Fig%%%%%%%%%%%%%%%%%%%%%%%%%%%%%
\begin{figure*}[t!]
  \centering
  \includegraphics[width=\linewidth]{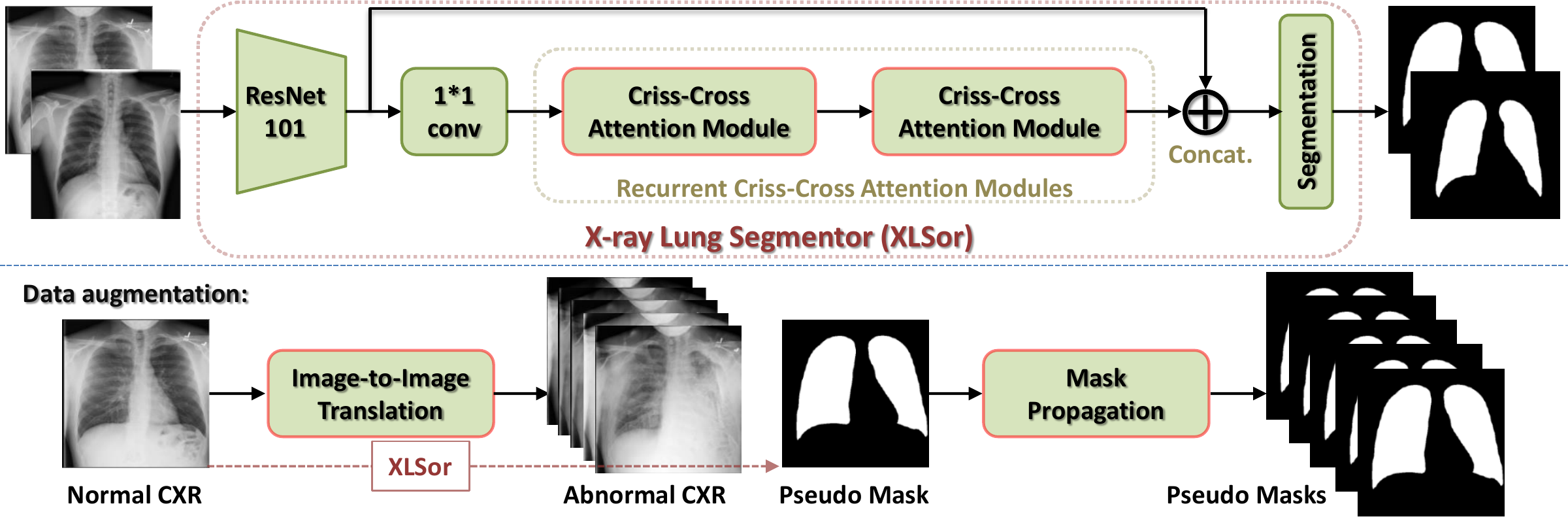}
  \caption{Framework  of the proposed X-ray lung segmentor (XLSor).}
  \label{fig:fw}
\end{figure*}
%%%%%%%%%%%%%%%%%%%%%%%%%%%%%%%%%%%%%%%%%%%%%%%%%%%%%%%%%%%%%

The overall XLSor framework is shown in Figure~\ref{fig:fw}. Given a training set $R$ with ground-truth masks, an initial lung segmentor is trained (see details in Sec.~\ref{ccnet}). Then, for an auxiliary external set, an image-to-image method MUNIT~\cite{munit} is used to synthesize abnormal CXRs from normal ones, so as to augment the training data and pseudo mask annotations (mask of normal CXR is obtained using the initial lung segmentor and propagated to its synthesized abnormal CXRs, see details in Sec.~\ref{munit}). The initial lung segmentor is updated using $R$ along with the augmented dataset $A$ with pseudo masks.

\subsection{Criss-Cross Attention based Network for Lung Segmentation}
\label{ccnet}

In preliminary experiments, we trained a U-Net model~\cite{U-net}, a widely used model in many applications of medical image segmentation, for lung segmentation. When testing it on the unseen abnormal CXRs, the segmentations are not very promising. That is because the features are extracted from local receptive fields and cannot well capture sufficient contextual information of lungs in U-Net. However, the rich and global contextual information of lungs and their surrounding regions is very important for lung segmentation.

Criss-cross Network (CCNet)~\cite{Huang2018CCNetCA} achieved state-of-the-art performance in semantic segmentation based on a novel criss-cross attention (CCA) module.
Inspired by this, we employ CCA to build a robust and accurate lung segmentor (named XLSor) on chest X-rays. The XLSor is constructed with a fully convolutional network and two CCA modules to capture long-range contextual information (see Figure~\ref{fig:fw} top). Specifically, we replace the last two down-sampling layers in the ImageNet pre-trained ResNet-101~\cite{resnet} with dilated convolution operation~\cite{deeplab}, resulting in an output stride of 8. The CCA module collects contextual
information in horizontal and vertical directions to enhance pixel-wise representative capability.
Recurrent criss-cross attention module can capture dense contextual information from all pixels by stacking two CCA modules with shared weights. CCA shares the similar idea of capturing global contextual information as the non-local neural network~\cite{nonlocal} but with much higher computational efficiency. Please refer to~\cite{Huang2018CCNetCA} for more details about the CCA module.
Therefore, the CCA based XLSor can generate clear lung boundaries for more accurate lung segmentation by considering the richer and global contextual information.

The mean square error loss function and the SGD with momentum of 0.9 and weight decay of 0.0005 are used to optimize the XLSor. The initial learning rate is 0.02 and updated using a poly learning rate policy where the initial learning rate is multiplied by $1-(\frac{iter}{max\_iter})^{0.9}$, where $iter$ is the number of current iterations and $max\_iter$ is the total number of iterations. The batch size is set as 4. The size of the input CXR is $512\times512$.

\subsection{Data Augmentation via Abnormal Chest X-Ray Pairs Construction}
\label{munit}
As discussed in Sec.~\ref{intro}, it is insufficient to train a robust lung segmentor using the existing datasets and mask annotations.  A simple solution is to enrich the training data, which has been widely used in deep learning. The traditional data augmentation means is to use a combination of affine transformations to manipulate the training data, \eg shifting, zooming in/out, rotation, flipping, etc, so as to generate new duplicate images for each input image. The contextual information in these generated images do not change very much. To solve these problems, we propose a data augmentation strategy using an image-to-image translation method~\cite{munit} to construct a large number of abnormal chest X-ray pairs without involving any human intervention, based on which a powerful model can be learned for robust and accurate lung segmentation on different challenging CXRs.

\begin{figure}[t!]
	\begin{minipage}[b]{1.0\linewidth}
		\centering
		\includegraphics[width=0.99\linewidth]{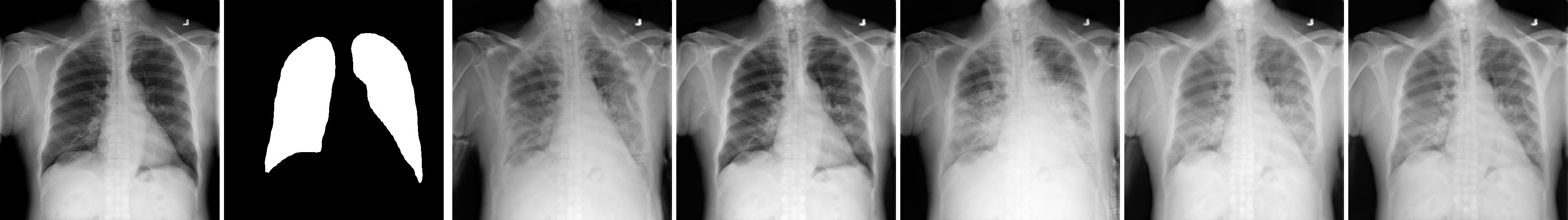} \\
		\vspace{0.05cm}
		\includegraphics[width=0.99\linewidth]{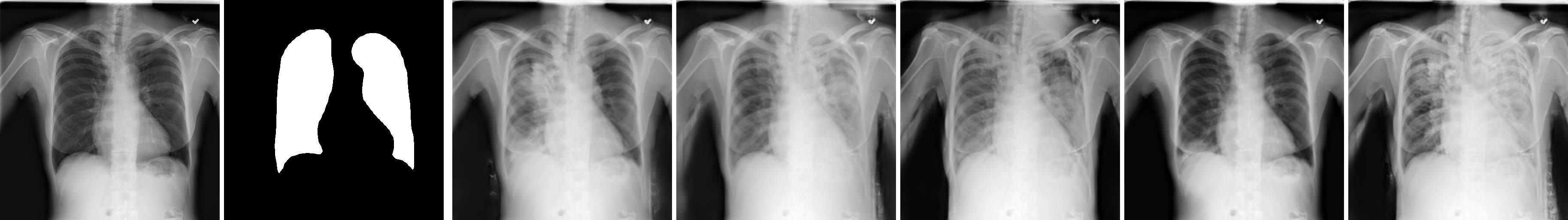} \\
		\vspace{0.05cm}
		\includegraphics[width=0.99\linewidth]{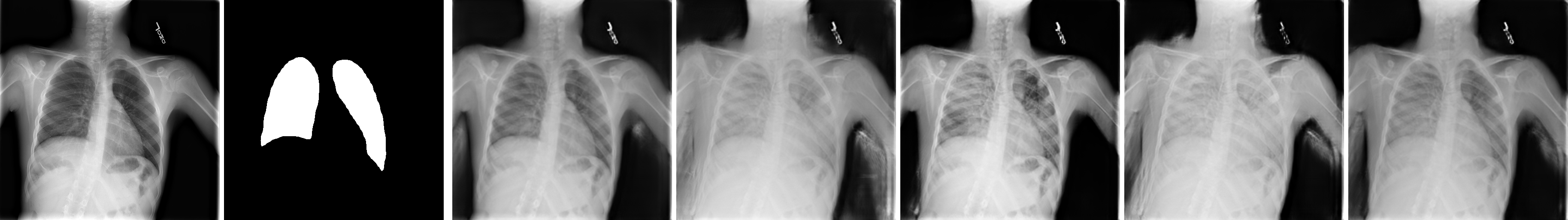} \\
	\end{minipage}
	\begin{minipage}[b]{0.136\linewidth}
		\centering
		\vspace{0.2cm}
		\centerline{(a)}\medskip
	\end{minipage}
	\begin{minipage}[b]{0.136\linewidth}
		\centering
		\vspace{0.2cm}
		\centerline{(b)}\medskip
	\end{minipage}
	\begin{minipage}[b]{0.136\linewidth}
		\centering
		\vspace{0.2cm}
		\centerline{(c)}\medskip
	\end{minipage}
	\begin{minipage}[b]{0.136\linewidth}
		\centering
		\vspace{0.2cm}
		\centerline{(d)}\medskip
	\end{minipage}
	\begin{minipage}[b]{0.136\linewidth}
		\centering
		\vspace{0.2cm}
		\centerline{(e)}\medskip
	\end{minipage}
	\begin{minipage}[b]{0.136\linewidth}
		\centering
		\vspace{0.2cm}
		\centerline{(f)}\medskip
	\end{minipage}
	\begin{minipage}[b]{0.136\linewidth}
		\centering
		\vspace{0.2cm}
		\centerline{(g)}\medskip
	\end{minipage}
	\caption{Three examples (rows) of the constructed abnormal CXR pairs. Given an unseen normal CXR (a), XLSor outputs a lung segmentation that is binarized with a threshold of 0.5 to get the lung mask (b) and MUNIT generates different abnormal CXRs (c-g). The lung mask (b) and the synthesized abnormal CXRs (c-g) form the constructed abnormal CXR pairs.}
	\label{fig:CXR_generation}
\end{figure}

To construct the pairs of abnormal CXR and its corresponding lung masks, there are two straightforward ways. One is to convert the abnormal CXRs into normal ones, and then compute the lung masks which serve as the ground truths for the abnormal CXRs. The other one is to convert the normal CXRs into abnormal ones, and then the lung masks segmented on the normal CXRs are considered as the ground truths of the abnormal ones. Here, we prefer the second way, since the lung regions in real normal CXRs are determined while the ones could be different for various generated normal CXRs in the first way. For the image-to-image translation task, \ie from normal CXRs to abnormal ones, a state-of-the-art method, \ie MUNIT \cite{munit}, is utilized in this work. MUNIT assumes that the image representation can be decomposed into a content code that is domain-invariant, and a style code that captures domain-specific properties. To translate an image to another domain, MUNIT recombines its content code with a random style code sampled from the style space of the target domain. Please refer to \cite{munit} for more details about MUNIT. In this work, we first train the MUNIT model using the default parameter configuration and the NIH chest X-Ray dataset~\cite{Wang_cvpr17}, from which 5,000 normal CXRs and 5,000 abnormal CXRs are randomly selected for training. Then, given a normal CXR (see Figure \ref{fig:CXR_generation}(a)), we use the trained MUNIT model to generate (or synthesize) a number of abnormal CXRs (see Figure \ref{fig:CXR_generation}(c)-(g)) by combining the content code of the normal CXR and different random style codes learned from the domain of abnormal CXRs. From Figure \ref{fig:CXR_generation}(c)-(g), we can see that the generated abnormal CXRs are radiorealistic. We also notice that the shape of lungs are distorted slightly in the generated abnormal CXRs sometimes. Therefore, the generated abnormalities are customized using the style codes and visually radiorealistic. At last, we use the initial XLSor model trained from the publicly available datasets to obtain the lung masks (see Figure \ref{fig:CXR_generation}(b)) of the given normal CXR, which are also considered as the pseudo masks of the generated abnormal CXRs (\ie mask propagation) to form the constructed abnormal CXR pairs (see Figure~\ref{fig:fw} bottom) for further training the XLSor model. We also iteratively conducted above processes and found that it is not helpful because the normal CXRs are easy to segment and the pseudo masks are good enough at the first iteration.

\section{Experiments}
\subsection{Datasets and Evaluation Criteria}
We evaluate the lung segmentation performance of the proposed XLSor using two publicly available datasets, \ie JSRT~\cite{JSRT} and Montgomery~\cite{TB_data}, and our own annotated dataset (named NIH). \textbf{JSRT} contains 247 CXRs, among which 154 have lung nodules and 93 have no lung nodule. \textbf{Montgomery} contains 138 CXRs, including 80 normal patients and 58 patients with manifested tuberculosis (TB). Both datasets provide pixel-wise lung mask annotations. We notice that the abnormal lung regions in these two datasets are mild. Only using such datasets for evaluation cannot well demonstrate the effectiveness and generalizability of the methods, since diseases can occasionally cause severe damages to the lungs. Therefore, we manually annotate the lung masks of 100 abnormal CXRs with various severity of lung diseases, which are selected from the NIH Chest X-Ray dataset~\cite{Wang_cvpr17} by excluding the samples used for MUNIT training. Here, we name the manually labeled set as \textbf{NIH}.

JSRT and Montgomery datasets are combined and randomly split into three subsets for both normal and abnormal CXRs, \ie training (70\%), validation (10\%) and testing (20\%). Specifically, the validation and testing sets include 37 and 78 CXRs, respectively. The remaining 280 CXRs serve as a training set for model training. The validation set is used for model selection, and the testing set and the NIH dataset are used for performance evaluation. Five criteria, \ie volumetric similarity (VS), averaged Hausdorff distance (AVD), Dice similarity coefficient (DICE), precision (PRE) and recall (REC) scores, are calculated pixel-wisely by a publicly available segmentation evaluation tool \cite{taha2015metrics} with threshold of 0.5 and used to evaluate the quantitative segmentation performance.

\subsection{Quantitative Results}
In this work, U-Net~\cite{U-net} is applied for performance comparisons to demonstrate the effectiveness of the criss-cross attention based XLSor. 
To validate the usefulness of adding the augmented samples for lung segmentation, we first use the proposed data augmentation strategy to generate four augmented training sets, denoted as $A^1$, $A^2$, $A^3$ and $A^4$, respectively. Here, $A^1$ contains 600 constructed pairs including 100 normal pairs and 500 abnormal pairs where five abnormal CXRs are synthesized from each normal CXR using MUNIT~\cite{munit}. $A^i$ $(i=2,3,4)$ contains all samples in $A^{i-1}$ and another new 600 constructed pairs.
We then train the XLSor and U-Net models for lung segmentation using six different training settings, \ie only using the real public training set (denoted $R$), using the real public training set and any augmented set $A^i$ $i=1,2,3,4$ (denoted $R+A^i$), and only using the augmented set $A^4$. 
To validate the effectiveness of CCA for segmentation performance improvement, we also train the XLSor model without CCA modules (denoted XLSor$^-$) and the U-Net model with CCA modules (denoted U-Net$^+$) using $R$ and $R+A^4$.
In each training setting, the same traditional data augmentation techniques (\eg scaling and flipping) are adopted. Finally, the five criteria are used to evaluate the performance of lung segmentation on the public testing set and NIH dataset, whose results are reported in Table \ref{tab:public}.

\begin{table}[t!]%htbp
\floatconts
  {tab:public}%
  {\caption{Lung segmentation results on the \textbf{public testing set} and \textbf{NIH dataset} using the proposed XLSor and U-Net with different training settings. Results showing mean with standard deviation. $\uparrow$: the larger the better. $\downarrow$: the smaller the better.}
  }%
  {\begin{tabular}{|c|c|c|c|c|c|}
  \hline
  \bfseries Method          & \bfseries REC$\uparrow$          & \bfseries PRE$\uparrow$         & \bfseries DICE$\uparrow$            & \bfseries AVD$\downarrow$         & \bfseries VS$\uparrow$         \\ \hline \hline
  \multicolumn{6}{|c|}{\textit{\textbf{Public testing set}}}             \\ \hline
XLSor$_{R}$        & 0.973$\pm$0.02  & \bfseries 0.979$\pm$0.02  & \bfseries 0.976$\pm$0.01  & 0.149$\pm$0.51  & \bfseries 0.992$\pm$0.01  \\
XLSor$_{R+A^1}$     & 0.973$\pm$0.02  & \bfseries 0.979$\pm$0.02  & \bfseries 0.976$\pm$0.01  & 0.152$\pm$0.52  & 0.991$\pm$0.01  \\ 
XLSor$_{R+A^2}$     & \bfseries 0.974$\pm$0.02  & 0.978$\pm$0.02  & \bfseries 0.976$\pm$0.01  & \bfseries 0.117$\pm$0.31  & 0.991$\pm$0.01  \\
XLSor$_{R+A^3}$     & 0.972$\pm$0.02  & \bfseries 0.979$\pm$0.02  & \bfseries 0.976$\pm$0.01  & 0.126$\pm$0.33  & 0.991$\pm$0.01  \\
XLSor$_{R+A^4}$     & \bfseries 0.974$\pm$0.02  & 0.977$\pm$0.02  & \bfseries 0.976$\pm$0.01  & 0.146$\pm$0.44  & 0.991$\pm$0.01  \\
XLSor$_{A^4}$       & 0.965$\pm$0.03  & \bfseries 0.979$\pm$0.02  & 0.972$\pm$0.02  & 0.162$\pm$0.36  & 0.989$\pm$0.01  \\ \hline
XLSor$_{R}^-$       & 0.973$\pm$0.02  & 0.978$\pm$0.02 & 0.975$\pm$0.01 & 0.151$\pm$0.53 & 0.991$\pm$0.01   \\
XLSor$_{R+A^4}^-$      & 0.972$\pm$0.02  & 0.978$\pm$0.02 & 0.976$\pm$0.01 & 0.148$\pm$0.47 & 0.991$\pm$0.01  \\ \hline

U-Net$_{R}$         & 0.976$\pm$0.02  & 0.968$\pm$0.03  & 0.972$\pm$0.02  & 0.198$\pm$0.56  & 0.988$\pm$0.02  \\
U-Net$_{R+A^1}$      & 0.973$\pm$0.02  & 0.976$\pm$0.02  & 0.974$\pm$0.01  & 0.162$\pm$0.54  & \bfseries 0.990$\pm$0.01  \\
U-Net$_{R+A^2}$      & \bfseries 0.977$\pm$0.02  & 0.973$\pm$0.02  & \bfseries 0.975$\pm$0.01  & 0.135$\pm$0.41  & 0.989$\pm$0.01  \\
U-Net$_{R+A^3}$      & 0.976$\pm$0.02  & 0.975$\pm$0.02  & \bfseries 0.975$\pm$0.01  & \bfseries 0.131$\pm$0.34  & \bfseries 0.990$\pm$0.01  \\
U-Net$_{R+A^4}$      & 0.973$\pm$0.02  & \bfseries 0.978$\pm$0.01  & \bfseries 0.975$\pm$0.01  & 0.152$\pm$0.46  & \bfseries 0.990$\pm$0.01  \\
U-Net$_{A^4}$        & 0.967$\pm$0.02  & 0.975$\pm$0.01  & 0.971$\pm$0.01  & 0.164$\pm$0.37  & 0.989$\pm$0.01  \\ \hline 
U-Net$_{R}^+$      &  0.976$\pm$0.02    &    0.970$\pm$0.03     &   0.972$\pm$0.02    &     0.191$\pm$0.54   &     0.988$\pm$0.02     \\
U-Net$_{R+A^4}^+$      &  0.975$\pm$0.02    &    0.977$\pm$0.01   &     0.975$\pm$0.01   &      0.130$\pm$0.33   &     0.990$\pm$0.01     \\ \hline \hline

\multicolumn{6}{|c|}{\textit{\textbf{NIH dataset}}}             \\ \hline
XLSor$_{R}$        & 0.966$\pm$0.02  & 0.927$\pm$0.09  & 0.943$\pm$0.05  & 0.669$\pm$1.64  & 0.966$\pm$0.05  \\
XLSor$_{R+A^1}$     & 0.958$\pm$0.03  & 0.973$\pm$0.02  & 0.965$\pm$0.02  & 0.172$\pm$0.26  & 0.985$\pm$0.01  \\
XLSor$_{R+A^2}$     & 0.962$\pm$0.02  & 0.980$\pm$0.01  & 0.971$\pm$0.01  & 0.097$\pm$0.08  & 0.989$\pm$0.01  \\
XLSor$_{R+A^3}$     & 0.967$\pm$0.02  & 0.978$\pm$0.02  & 0.973$\pm$0.01  & 0.089$\pm$0.07  & 0.990$\pm$0.01  \\
XLSor$_{R+A^4}$     & \bfseries 0.974$\pm$0.01  & 0.976$\pm$0.01  & \bfseries 0.975$\pm$0.01  & \bfseries 0.078$\pm$0.06  & \bfseries 0.993$\pm$0.01  \\
XLSor$_{A^4}$       & 0.964$\pm$0.02  & \bfseries 0.983$\pm$0.01  & 0.973$\pm$0.01  & 0.098$\pm$0.13  & 0.988$\pm$0.01  \\ \hline

XLSor$_{R}^-$      & 0.965$\pm$0.03 & 0.902$\pm$0.10 & 0.929$\pm$0.06 & 0.952$\pm$1.81 & 0.955$\pm$0.06  \\
XLSor$_{R+A^4}^-$      & 0.965$\pm$0.02 & 0.981$\pm$0.01 & 0.967$\pm$0.01 & 0.093$\pm$0.10 & 0.990$\pm$0.01  \\ \hline

U-Net$_{R}$         & 0.938$\pm$0.07  & 0.761$\pm$0.20  & 0.823$\pm$0.16  & 5.231$\pm$9.02  & 0.869$\pm$0.15  \\
U-Net$_{R+A^1}$      & 0.926$\pm$0.05  & \bfseries 0.960$\pm$0.03  & 0.942$\pm$0.03  & 0.832$\pm$1.29  & 0.971$\pm$0.02  \\
U-Net$_{R+A^2}$      & 0.947$\pm$0.04  & 0.950$\pm$0.04  & 0.948$\pm$0.03  & 0.500$\pm$1.03  & 0.981$\pm$0.02  \\
U-Net$_{R+A^3}$      & 0.950$\pm$0.03  & 0.954$\pm$0.03  & 0.951$\pm$0.02  & 0.393$\pm$0.58  & 0.983$\pm$0.02  \\
U-Net$_{R+A^4}$      & 0.943$\pm$0.04  & 0.958$\pm$0.03  & 0.950$\pm$0.03  & 0.454$\pm$0.73  & 0.982$\pm$0.02  \\
U-Net$_{A^4}$        & \bfseries0.952$\pm$0.03  & 0.959$\pm$0.03  & \bfseries 0.955$\pm$0.02  & \bfseries 0.315$\pm$0.47  & \bfseries 0.983$\pm$0.02  \\ \hline
U-Net$_{R}^+$      & 0.929$\pm$0.07    &    0.804$\pm$0.20     &   0.842$\pm$0.14   &      4.782$\pm$8.05    &    0.895$\pm$0.14   \\
U-Net$_{R+A^4}^+$      & 0.956$\pm$0.03   &     0.969$\pm$0.02   &     0.962$\pm$0.02    &     0.262$\pm$0.54   &     0.985$\pm$0.02     \\ \hline
  \end{tabular}}
\end{table}

\begin{figure}[t!]
\tiny{
    \begin{minipage}[b]{0.12\linewidth}
		\centering
% 		\vspace{0.2cm}
		\centerline{CXR}
	\end{minipage}
	\begin{minipage}[b]{0.12\linewidth}
		\centering
% 		\vspace{0.2cm}
		\centerline{Mask}
	\end{minipage}
	\begin{minipage}[b]{0.12\linewidth}
		\centering
% 		\vspace{0.2cm}
		\centerline{U-Net$_{R}$}
	\end{minipage}
	\begin{minipage}[b]{0.12\linewidth}
		\centering
% 		\vspace{0.2cm}
		\centerline{U-Net$_{A^4}$}
	\end{minipage}
	\begin{minipage}[b]{0.11\linewidth}
		\centering
% 		\vspace{0.2cm}
		\centerline{U-Net$_{R+A^4}$}
	\end{minipage}
	\begin{minipage}[b]{0.12\linewidth}
		\centering
% 		\vspace{0.2cm}
		\centerline{XLSor$_{R}$}
	\end{minipage}
	\begin{minipage}[b]{0.11\linewidth}
		\centering
% 		\vspace{0.2cm}
		\centerline{XLSor$_{A^4}$}
	\end{minipage}
	\begin{minipage}[b]{0.12\linewidth}
		\centering
% 		\vspace{0.2cm}
		\centerline{XLSor$_{R+A^4}$}
	\end{minipage}}
	\begin{minipage}[b]{1.0\linewidth}
		\centering
		\includegraphics[width=0.99\linewidth]{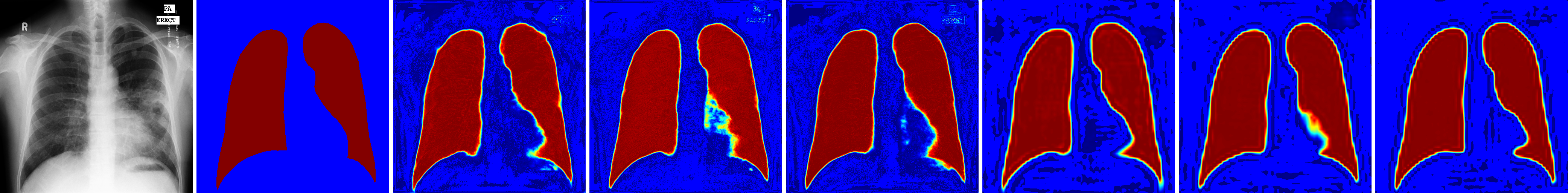} \\
		\vspace{0.05cm}
		\includegraphics[width=0.99\linewidth]{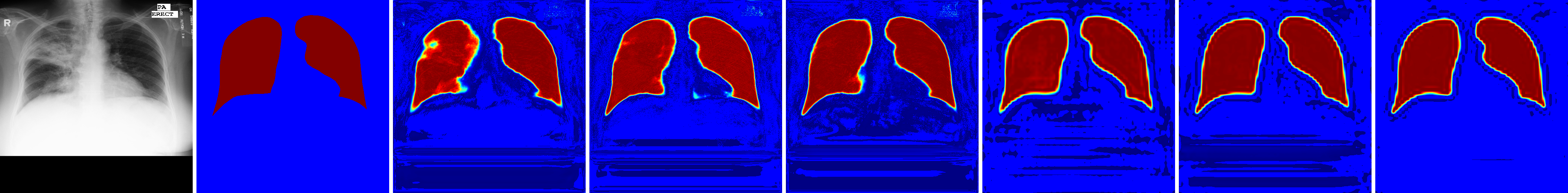} \\
		\vspace{0.05cm}
		\includegraphics[width=0.99\linewidth]{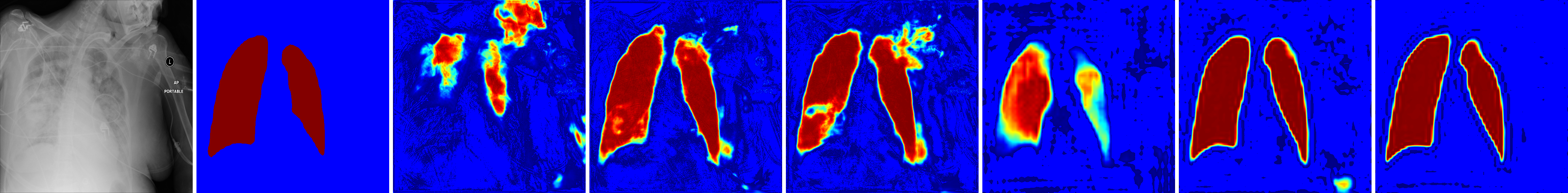} \\
		\vspace{0.05cm}
		\includegraphics[width=0.99\linewidth]{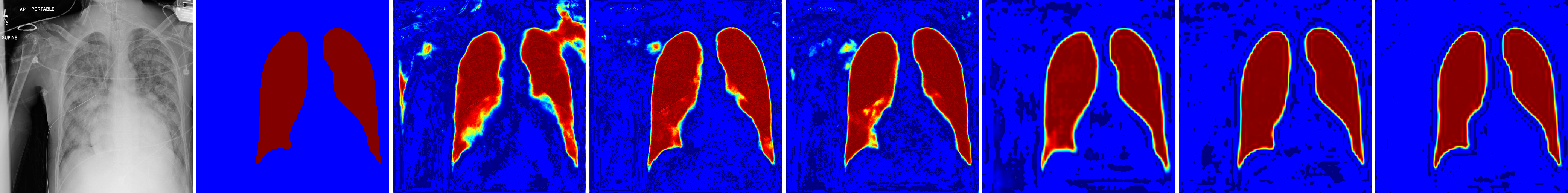} \\
	\end{minipage}
	\normalsize{
	\begin{minipage}[b]{0.122\linewidth}
		\centering
		\vspace{0.2cm}
		\centerline{(a)}\medskip
	\end{minipage}
	\begin{minipage}[b]{0.12\linewidth}
		\centering
		\vspace{0.2cm}
		\centerline{(b)}\medskip
	\end{minipage}
	\begin{minipage}[b]{0.12\linewidth}
		\centering
		\vspace{0.2cm}
		\centerline{(c)}\medskip
	\end{minipage}
	\begin{minipage}[b]{0.12\linewidth}
		\centering
		\vspace{0.2cm}
		\centerline{(d)}\medskip
	\end{minipage}
	\begin{minipage}[b]{0.115\linewidth}
		\centering
		\vspace{0.2cm}
		\centerline{(e)}\medskip
	\end{minipage}
	\begin{minipage}[b]{0.12\linewidth}
		\centering
		\vspace{0.2cm}
		\centerline{(f)}\medskip
	\end{minipage}
	\begin{minipage}[b]{0.12\linewidth}
		\centering
		\vspace{0.2cm}
		\centerline{(g)}\medskip
	\end{minipage}
	\begin{minipage}[b]{0.115\linewidth}
		\centering
		\vspace{0.2cm}
		\centerline{(h)}\medskip
	\end{minipage}}
	\caption{Four examples (rows) of lung segmentation results produced by XLSor and U-Net trained using $R$, $A^4$ and $R+A^4$. Here, the results are given as the probability maps directly outputted by the models, which can be binarized with a threshold of 0.5 to get the binary lung masks for performance evaluation. The first two rows are from the public testing set and the last two rows are from the NIH dataset. To better visualize the differences between lung segmentation results and ground truths, we colorize them with pseudo-colors. Better viewed in color.}
	\label{fig:lung_segmentation}
\end{figure}

From Table \ref{tab:public}, we can see that 1) the proposed XLSor gets better results than U-Net on both the simple public testing set and the difficult NIH dataset. Especially, the performance of XLSor$_{R}$ is much better than the one of U-Net$_{R}$ on the NIH dataset (\eg improving the Dice score about 12\%), meaning that the proposed XLSor is able to work much better than U-Net on the unseen CXRs whose data distribution is much different from the training data. This demonstrates that the proposed XLSor based on the criss-cross attention module can well learn the global contextual information of lung regions and strong discriminative features to distinguish the lung regions from their surrounding structures regardless of the CXRs' properties. 2) When adding the augmented samples for model training, the performance is improved, \ie XLSor$_{R+A^i}$ (or U-Net$_{R+A^i}$) gets better results than XLSor$_{R}$ (or U-Net$_{R}$), suggesting the effectiveness of our data augmentation technique for lung segmentation performance improvement. Through experiments, we find that the performance remains stable when adding more augmented samples than $A^4$. 3) When only using the augmented samples for model training, both XLSor and U-Net still get very promising performance on the public testing set and the NIH dataset (see the results of XLSor$_{A^4}$ and U-Net$_{A^4}$ in Table \ref{tab:public}), suggesting that the generated abnormal CXRs are radiorealistic and the pseudo lung masks effectively supervise the learning processes for lung segmentation. 4) The results by all models are quite similar in the public testing set, that is because the testing CXRs are all (near-)normal and the lung segmentation task is relatively easy. 5) U-Net obtains worse performance on NIH dataset than the public testing set, meaning that the CXRs in the NIH dataset are more complex and difficult than the ones in the public testing set. But XLSor can get comparable and good results on both datasets, suggesting that the proposed XLSor is robust and powerful for lung segmentation in different scenarios. 6) XLSor/U-Net$^+$ achieves better results than XLSor$^-$/U-Net (especially, on the NIH dataset), suggesting that using CCA modules can make the model learn the global contextual information of lung regions better and extract more powerful discriminative features for performance improvement. All results quantitatively demonstrate the effectiveness and generalizability of the proposed XLSor for lung segmentation on various CXRs.

\subsection{Qualitative Results}
Figure \ref{fig:lung_segmentation} shows four qualitative lung segmentation results produced by the models (\ie XLSor and U-Net) trained with the following settings: $R$, $A^4$ and $R+A^4$. Compared with U-Net, the lung segmentation results produced by the proposed XLSor are much closer to the ground truths in various challenging scenarios. To be specific, 1) the proposed XLSor not only highlights the correct lung regions clearly, but also well suppresses the probabilities of background regions, so as to produce the segmentation results
with higher contrast between lung regions and background than U-Net. 2) With the help of the criss-cross attention module that considers sufficient contextual information, the proposed XLSor is able to output the lung segmentations with clear boundaries and consistent probabilities, even when the model is trained and tested on CXRs with different distribution of abnormalities. 3) With the augmented samples for training, the qualities of lung segmentations are improved. These intuitively demonstrate the effectiveness of the proposed XLSor and the usefulness of the proposed data augmentation strategy for lung segmentation on chest X-rays. 

\section{Conclusions and Future Work}
In this paper, we propose a robust and accurate lung segmentor based on a criss-cross attention network and a customized radiorealistic abnormalities generation technique for data augmentation. Experiments showed that the proposed framework was able to capture rich contextual information from both original and radiorealistic synthesized CXRs to adapt to more challenging images, resulting in much better segmentation, especially in unseen abnormal CXRs. Future work includes segmenting more organs and integrating with more downstream tasks such as disease classification and detection to provide comprehensive and accurate computer-aided detection on CXR images, \eg performing segmentation and classification simultaneously by training different MUNIT models for individual diseases and using them to generate abnormalities accordingly in categories.

\midlacknowledgments{This research was supported by the Intramural Research Program of the National Institutes of Health Clinical Center and by the Ping An Insurance Company through a Cooperative Research and Development Agreement. We thank Nvidia for GPU card donation.}

\bibliography{tang19}

\begin{thebibliography}{31}
\providecommand{\natexlab}[1]{#1}
\providecommand{\url}[1]{\texttt{#1}}
\expandafter\ifx\csname urlstyle\endcsname\relax
  \providecommand{\doi}[1]{doi: #1}\else
  \providecommand{\doi}{doi: \begingroup \urlstyle{rm}\Url}\fi

\bibitem[Cai et~al.(2018)Cai, Tang, Lu, Harrison, Yan, Xiao, Yang, and
  Summers]{cai2018accurate}
J.~Cai, Y.~Tang, L.~Lu, A.~P Harrison, K.~Yan, J.~Xiao, L.~Yang, and R.~M.
  Summers.
\newblock Accurate weakly-supervised deep lesion segmentation using large-scale
  clinical annotations: Slice-propagated 3{D} mask generation from 2{D}
  {RECIST}.
\newblock In \emph{MICCAI}, 2018.

\bibitem[Candemir et~al.(2014)Candemir, Jaeger, Palaniappan, Musco, Singh, Xue,
  Karargyris, Antani, Thoma, and McDonald]{Candemir_TMI14}
S.~Candemir, S.~Jaeger, K.~Palaniappan, J.~P. Musco, R.~K. Singh, Z.~Xue,
  A.~Karargyris, S.~Antani, G.~Thoma, and C.~J. McDonald.
\newblock Lung segmentation in chest radiographs using anatomical atlases with
  nonrigid registration.
\newblock \emph{IEEE TMI}, 33\penalty0 (2):\penalty0 577--590, 2014.

\bibitem[Chen et~al.(2018)Chen, Bentley, Mori, Misawa, Fujiwara, and
  Rueckert]{DRINet}
L.~Chen, P.~Bentley, K.~Mori, K.~Misawa, M.~Fujiwara, and D.~Rueckert.
\newblock Drinet for medical image segmentation.
\newblock \emph{IEEE TMI}, 37\penalty0 (11):\penalty0 2453--2462, 2018.

\bibitem[Chen et~al.(2015)Chen, Papandreou, Kokkinos, Murphy, and
  Yuille]{deeplab}
L.~C. Chen, G.~Papandreou, I.~Kokkinos, K.~Murphy, and A.~L. Yuille.
\newblock Semantic image segmentation with deep convolutional nets and fully
  connected {CRF}s.
\newblock In \emph{ICLR}, 2015.

\bibitem[Dai et~al.(2017)Dai, Doyle, Liang, Zhang, Dong, Li, and
  Xing]{dai17scan}
W.~Dai, J.~Doyle, X.~Liang, H.~Zhang, N.~Dong, Y.~Li, and E.~P. Xing.
\newblock Scan: Structure correcting adversarial network for chest x-rays organ
  segmentation.
\newblock \emph{arXiv preprint arXiv:1703.08770}, 2017.

\bibitem[El-Baz et~al.(2016)El-Baz, Jiang, and Suri]{el2016biomedical}
A.~El-Baz, X.~Jiang, and J.S. Suri.
\newblock \emph{Biomedical Image Segmentation: Advances and Trends}.
\newblock CRC Press, Taylor \& Francis Group, 2016.
\newblock ISBN 9781482258554.

\bibitem[Goodfellow et~al.(2014)Goodfellow, Pouget-Abadie, Mirza, Xu,
  Warde-Farley, Ozair, Courville, and Bengio]{GAN}
I.~Goodfellow, J.~Pouget-Abadie, M.~Mirza, B.~Xu, D.~Warde-Farley, S.~Ozair,
  A.~Courville, and Y.~Bengio.
\newblock Generative adversarial nets.
\newblock In \emph{NeurIPS}. 2014.

\bibitem[He et~al.(2016)He, Zhang, Ren, and Sun]{resnet}
K.~He, X.~Zhang, S.~Ren, and J.~Sun.
\newblock Deep residual learning for image recognition.
\newblock In \emph{CVPR}, 2016.

\bibitem[Huang et~al.(2018{\natexlab{a}})Huang, Liu, Belongie, and
  Kautz]{munit}
X.~Huang, M.Y. Liu, S.~Belongie, and J.~Kautz.
\newblock Multimodal unsupervised image-to-image translation.
\newblock In \emph{ECCV}, 2018{\natexlab{a}}.

\bibitem[Huang et~al.(2018{\natexlab{b}})Huang, Wang, Huang, Huang, Wei, and
  Liu]{Huang2018CCNetCA}
Z.~Huang, X.~Wang, L.~Huang, C.~Huang, Y.~Wei, and W.~Liu.
\newblock {CCN}et: Criss-cross attention for semantic segmentation.
\newblock \emph{arXiv preprint arXiv:1811.11721}, 2018{\natexlab{b}}.

\bibitem[Jaeger et~al.(2014)Jaeger, Candemir, Antani, Lu, and Thoma]{TB_data}
S.~Jaeger, S.~Candemir, Y.~X. Antani, S.and~Wang, P.~X. Lu, and G.~Thoma.
\newblock Two public chest {X}-ray datasets for computer-aided screening of
  pulmonary diseases.
\newblock \emph{QIMS}, 4\penalty0 (6):\penalty0 475--477, 2014.

\bibitem[Jin et~al.(2018)Jin, Xu, Tang, Harrison, and Mollura]{jin2018ct}
D.~Jin, Z.~Xu, Y.~Tang, A.~P. Harrison, and D.~J. Mollura.
\newblock {CT}-realistic lung nodule simulation from 3{D} conditional
  generative adversarial networks for robust lung segmentation.
\newblock In \emph{MICCAI}, 2018.

\bibitem[Li et~al.(2001)Li, Zheng, Kallergi, and Clark]{LI2001629}
L.~Li, Y.~Zheng, M.~Kallergi, and R.~A. Clark.
\newblock Improved method for automatic identification of lung regions on chest
  radiographs.
\newblock \emph{AR}, 8\penalty0 (7):\penalty0 629 -- 638, 2001.

\bibitem[Ronneberger et~al.(2015)Ronneberger, Fischer, and Brox]{U-net}
O.~Ronneberger, P.~Fischer, and T.~Brox.
\newblock U-net: Convolutional networks for biomedical image segmentation.
\newblock In \emph{MICCAI}, 2015.

\bibitem[Shelhamer et~al.(2017)Shelhamer, Long, and Darrell]{FCN}
E.~Shelhamer, J.~Long, and T.~Darrell.
\newblock Fully convolutional networks for semantic segmentation.
\newblock \emph{IEEE TPAMI}, 39\penalty0 (4):\penalty0 640--651, 2017.

\bibitem[Shin et~al.(2016)Shin, Roberts, Lu, Demner-Fushman, Yao, and
  Summers]{Shin_2016_CVPR}
H.~C. Shin, K.~Roberts, L.~Lu, D.~Demner-Fushman, J.~Yao, and R.~M. Summers.
\newblock Learning to read chest {X}-rays: Recurrent neural cascade model for
  automated image annotation.
\newblock In \emph{CVPR}, 2016.

\bibitem[Shiraishi et~al.(2000)Shiraishi, Katsuragawa, Ikezoe, Matsumoto,
  Kobayashi, Komatsu, Matsui, Fujita, Kodera, and Doi]{JSRT}
J.~Shiraishi, S.~Katsuragawa, J.~Ikezoe, T.~Matsumoto, T.~Kobayashi,
  K.~Komatsu, M.~Matsui, H.~Fujita, Y.~Kodera, and K.~Doi.
\newblock Development of a digital image database for chest radiographs with
  and without a lung nodule.
\newblock \emph{AJR}, 174\penalty0 (1):\penalty0 71--74, 2000.

\bibitem[Taha and Hanbury(2015)]{taha2015metrics}
A.~A. Taha and A.~Hanbury.
\newblock Metrics for evaluating 3d medical image segmentation: analysis,
  selection, and tool.
\newblock \emph{BMC MI}, 15\penalty0 (1):\penalty0 29, 2015.

\bibitem[Tang et~al.(2018{\natexlab{a}})Tang, Cai, Lu, Harrison, Yan, Xiao,
  Yang, and Summers]{tang2018ct}
Y.~Tang, J.~Cai, L.~Lu, A.~P. Harrison, K.~Yan, J.~Xiao, L.~Yang, and R.~M.
  Summers.
\newblock {CT} image enhancement using stacked generative adversarial networks
  and transfer learning for lesion segmentation improvement.
\newblock In \emph{MLMI}, 2018{\natexlab{a}}.

\bibitem[Tang et~al.(2018{\natexlab{b}})Tang, Harrison, Bagheri, Xiao, and
  Summers]{tang2018semi}
Y.~Tang, A.~P. Harrison, M.~Bagheri, J.~Xiao, and R.~M. Summers.
\newblock Semi-automatic {RECIST} labeling on {CT} scans with cascaded
  convolutional neural networks.
\newblock In \emph{MICCAI}, 2018{\natexlab{b}}.

\bibitem[Tang et~al.(2018{\natexlab{c}})Tang, Wang, Harrison, Lu, Xiao, and
  Summers]{Tang_MLMI}
Y.~Tang, X.~Wang, A.~P. Harrison, L.~Lu, J.~Xiao, and R.~M. Summers.
\newblock Attention-guided curriculum learning for weakly supervised
  classification and localization of thoracic diseases on chest radiographs.
\newblock In \emph{MLMI}, 2018{\natexlab{c}}.

\bibitem[Tang et~al.(2019{\natexlab{a}})Tang, Oh, Tang, Xiao, and
  Summers]{tang2019ct}
Y.~B. Tang, S.~Oh, Y.~X. Tang, J.~Xiao, and R.~M. Summers.
\newblock {CT}-realistic data augmentation using generative adversarial network
  for robust lymph node segmentation.
\newblock In \emph{Medical Imaging: CAD}, 2019{\natexlab{a}}.

\bibitem[Tang et~al.(2019{\natexlab{b}})Tang, Yan, Tang, Liu, Xiao, and
  Summers]{tang2019uldor}
Y.~B. Tang, K.~Yan, Y.~X. Tang, J.~Liu, J.~Xiao, and R.~M. Summers.
\newblock {ULD}or: A universal lesion detector for {CT} scans with pseudo masks
  and hard negative example mining.
\newblock In \emph{ISBI}, 2019{\natexlab{b}}.

\bibitem[Tang et~al.(2019{\natexlab{c}})Tang, Tang, Han, Xiao, and
  Summers]{tang2019abnormal}
Y.~X. Tang, Y.~B. Tang, M.~Han, J.~Xiao, and R.~M. Summers.
\newblock Abnormal chest {X}-ray identification with generative adversarial
  one-class classifier.
\newblock In \emph{ISBI}, 2019{\natexlab{c}}.

\bibitem[Tang et~al.(2019{\natexlab{d}})Tang, Tang, Han, Xiao, and
  Summers]{tang2019deep}
Y.~X. Tang, Y.~B. Tang, M.~Han, J.~Xiao, and R.~M. Summers.
\newblock Deep adversarial one-class learning for normal and abnormal chest
  radiograph classification.
\newblock In \emph{Medical Imaging: CAD}, 2019{\natexlab{d}}.

\bibitem[Wang et~al.(2017)Wang, Peng, Lu, Lu, Bagheri, and
  Summers]{Wang_cvpr17}
X.~Wang, Y.~Peng, L.~Lu, Z.~Lu, M.~Bagheri, and R.~M. Summers.
\newblock Chest{X}-ray8: Hospital-scale chest x-ray database and benchmarks on
  weakly-supervised classification and localization of common thorax diseases.
\newblock In \emph{CVPR}, 2017.

\bibitem[Wang et~al.(2018{\natexlab{a}})Wang, Girshick, Gupta, and
  He]{nonlocal}
X.~Wang, R.~Girshick, A.~Gupta, and K.~He.
\newblock Non-local neural networks.
\newblock In \emph{CVPR}, 2018{\natexlab{a}}.

\bibitem[Wang et~al.(2018{\natexlab{b}})Wang, Peng, Lu, Lu, and
  Summers]{Wang_cvpr18}
X.~Wang, Y.~Peng, L.~Lu, Z.~Lu, and R.~M. Summers.
\newblock Tienet: Text-image embedding network for common thorax disease
  classification and reporting in chest {X}-rays.
\newblock In \emph{CVPR}, 2018{\natexlab{b}}.

\bibitem[Xu et~al.(2012)Xu, Mandal, Long, Cheng, and Basu]{XU2012452}
T.~Xu, M.~Mandal, R.~Long, I.~Cheng, and A.~Basu.
\newblock An edge-region force guided active shape approach for automatic lung
  field detection in chest radiographs.
\newblock \emph{CMIG}, 36\penalty0 (6):\penalty0 452 -- 463, 2012.

\bibitem[Yan et~al.(2018)Yan, Wang, Lu, Zhang, Harrison, Bagheri, and
  Summers]{yan2018deep}
K.~Yan, X.~Wang, L.~Lu, L.~Zhang, A.~P. Harrison, M.~Bagheri, and R.~M.
  Summers.
\newblock Deep lesion graphs in the wild: relationship learning and
  organization of significant radiology image findings in a diverse large-scale
  lesion database.
\newblock In \emph{CVPR}, 2018.

\bibitem[Yan et~al.(2019)Yan, Peng, Lu, and Summers]{yan2019fine}
K.~Yan, Y.~Peng, Z.~Lu, and R.~M. Summers.
\newblock Fine-grained lesion annotation in {CT} images with knowledge mined
  from radiology reports.
\newblock In \emph{CVPR}, 2019.

\end{thebibliography}
\end{document}